\documentclass[table]{ai2style/ai2}

\usepackage{amssymb} % Loaded in commands.tex

\usepackage{multirow}
\usepackage{bigdelim}
\usepackage{todonotes}
\usepackage{longtable}
\usepackage{tabularray}
\usepackage{wrapfig}
\usepackage[most]{tcolorbox} 
\usepackage{url}
\usepackage{xurl}
\usepackage{xspace}
\usepackage{svg}
\usepackage[absolute]{textpos}
\usepackage{fdsymbol}
\usepackage[utf8]{inputenc}  % Not needed with XeTeX
\usepackage[T1]{fontenc}     % Not needed with XeTeX (fontspec handles this)
\usepackage{booktabs}
\usepackage{amsfonts}
\usepackage{nicefrac}
\usepackage[table]{xcolor}
\usepackage{amsmath}
\usepackage{csquotes}
\usepackage{siunitx}
\usepackage{graphicx}
\usepackage{arydshln}
\usepackage{enumitem}
\usepackage{soul}
\usepackage{adjustbox}
\usepackage{pifont}
\usepackage{caption}
\usepackage{makecell}
\usepackage{subcaption}
\usepackage{listings}
\usepackage{tikz}
\usepackage{nicematrix}

\usepackage{svg}
\usepackage{lineno}
\usepackage{float}
\usepackage{pgfplots}
\usepackage{pgfplotstable}
\usepackage{tikz-layers}

\usetikzlibrary{pgfplots.groupplots}
\pgfplotsset{compat=1.3}
\pgfplotsset{
  layers/axis lines on top/.define layer set={
    axis background,
    axis grid,
    axis ticks,
    axis tick labels,
    pre main,
    main,
    axis lines,
    axis descriptions,
    axis foreground,
  }{/pgfplots/layers/standard},
}
\usetikzlibrary{patterns}
\usetikzlibrary{shapes.geometric}

\definecolor{maroon}{HTML}{F26035}
\definecolor{yellow}{HTML}{FDBC42}
\definecolor{lavender}{HTML}{734f96}
\definecolor{darkergrey}{HTML}{444444}
\definecolor{midgrey}{HTML}{e6eded}
\definecolor{ai2pink}{HTML}{f0529c}
\definecolor{ai2midpink}{HTML}{fad3e5}
\definecolor{ai2lightpink}{HTML}{fbecf3}
\definecolor{ai2midwhite}{HTML}{f2e5d9}
\definecolor{ai2offwhite}{HTML}{fbf4ee}
\definecolor{ai2green}{HTML}{0fcb8c}
\definecolor{ai2lightgreen}{HTML}{e7f9f3}
\definecolor{ai2darkgreen}{HTML}{105257}
\definecolor{ai2purple}{HTML}{B932EB}
\definecolor{ai2lightpurple}{HTML}{f7e8fc}
\definecolor{neutralEight}{HTML}{343434}
\definecolor{neutralFive}{HTML}{838383}
\definecolor{neutralThree}{HTML}{bebebe}
\definecolor{neutralOne}{HTML}{dedede}
\definecolor{lightgrey}{HTML}{fafcfc}

\definecolor{SwordOrange}{HTML}{ff8351}
\definecolor{SwordBlueComplentarySwordOrange}{HTML}{51cdff}
\definecolor{SwordBlue}{HTML}{5993ea}
\definecolor{SwordSilver}{HTML}{a4aab6}
\definecolor{SwordTan}{HTML}{dacdc3}
\definecolor{SwordNoir}{HTML}{20222c}
\definecolor{SwordRed}{HTML}{ff8283}
\definecolor{SwordYellow}{HTML}{ffda51}
\definecolor{SwordSquash}{HTML}{00c487}
\definecolor{SwordPink}{HTML}{ed75ff}

\definecolor{linkcolor}{RGB}{0, 0, 128}
\hypersetup{
     colorlinks   = true,
     citecolor    = linkcolor,
     linkcolor    = linkcolor,
     urlcolor     = linkcolor,
}

\usepackage{CJKutf8}

\usepackage{hyperref}
\usepackage{url}

\usepackage{amssymb}
\usepackage{amsmath}
\usepackage{amsthm}
\usepackage{amsfonts}
\usepackage{bm}
\usepackage{nicefrac}
\usepackage{dsfont}
\usepackage{fancyvrb}
\usepackage{fvextra}
\usepackage{listings}
\usepackage{xcolor}
\usepackage{tabularx}% added for table design
\usepackage[all]{nowidow}
\usepackage{xcolor}
\usepackage{colortbl}
\usepackage{changepage}  % For adjustwidth environment

\usepackage{booktabs}
\usepackage{multirow}
\usepackage{makecell}
\usepackage{arydshln}

\usepackage{comment}
\usepackage{xspace}
\usepackage{soul}

\usepackage{enumitem}
\usepackage[normalem]{ulem}
\setlist[itemize,enumerate]{leftmargin=*}

\makeatletter
\def\adl@drawiv#1#2#3{%
        \hskip.5\tabcolsep
        \xleaders#3{#2.5\@tempdimb #1{1}#2.5\@tempdimb}%
                #2\z@ plus1fil minus1fil\relax
        \hskip.5\tabcolsep}
\newcommand{\cdashlinelr}[1]{%
  \noalign{\vskip 2pt
           \global\let\@dashdrawstore\adl@draw
           \global\let\adl@draw\adl@drawiv}
  \cdashline{#1}[.4pt/2pt]
  \noalign{\global\let\adl@draw\@dashdrawstore
           \vskip 2pt}}
\makeatother

\definecolor{light-orange}{HTML}{fee9d4}
\definecolor{light-green}{HTML}{d8f0d3}
\definecolor{light-blue}{HTML}{dae8f5}
\definecolor{set10-red}{HTML}{e41a1c}
\definecolor{set10-blue}{HTML}{377eb8}
\definecolor{set10-green}{HTML}{4daf4a}

\definecolor{CustomBlue}{RGB}{57,83,191}
\definecolor{CustomRed}{HTML}{a75151}
\definecolor{DarkGreenOne}{RGB}{106,168,79}

\definecolor{SwordOrange}{HTML}{ff8351}
\definecolor{SwordBlueComplentarySwordOrange}{HTML}{51cdff}
\definecolor{SwordBlue}{HTML}{5993ea}
\definecolor{SwordSilver}{HTML}{a4aab6}
\definecolor{SwordTan}{HTML}{dacdc3}
\definecolor{SwordNoir}{HTML}{20222c}
\definecolor{SwordRed}{HTML}{ff8283}
\definecolor{SwordYellow}{HTML}{ffda51}
\definecolor{SwordSquash}{HTML}{00c487}
\definecolor{SwordPink}{HTML}{ed75ff}

\definecolor{QwenPurple}{HTML}{6349ea}
\definecolor{GeminiBlue}{HTML}{4185f4}
\definecolor{OpenAIGreen}{HTML}{1fa681}
\definecolor{AnthropicTan}{HTML}{d5a583}
\definecolor{ZaiAsh}{HTML}{282828}
\usepackage[most]{tcolorbox}

\newtcbox{\clustertab}[1]{on line, box align=base, colback={#1},colframe={#1},size=fbox,arc=2pt,top=-1.5pt, bottom=-1.5pt, left=-1.5pt, right=-1.5pt, boxrule=0pt, enlarge left by=1pt}

\usepackage{pifont}% http://ctan.org/pkg/pifont
\newcommand{\sword}{}

\title[Arbor: A Framework for Reliable Navigation of Critical Conversation Flows]{Arbor: A Framework for Reliable Navigation of Critical Conversation Flows}

\authorOne[\sword]{Luís Silva}
\authorOne[\sword]{Diogo Gonçalves}
\authorOne[\sword]{Catarina Farinha}
\authorOne[\sword]{Clara Matos}  
\lastAuthorOne[\sword]{Luís Ungaro}

\affiliation[\sword]{Sword Health}

\abstract{
Large language models struggle to maintain strict adherence to structured workflows in high-stakes domains such as healthcare triage. Monolithic approaches that encode entire decision structures within a single prompt are prone to instruction-following degradation as prompt length increases, including lost-in-the-middle effects and context window overflow. To address this gap, we present Arbor, a framework that decomposes decision tree navigation into specialized, node-level tasks. Decision trees are standardized into an edge-list representation and stored for dynamic retrieval. At runtime, a directed acyclic graph (DAG)-based orchestration mechanism iteratively retrieves only the outgoing edges of the current node, evaluates valid transitions via a dedicated LLM call, and delegates response generation to a separate inference step. The framework is agnostic to the underlying decision logic and model provider. Evaluated against single-prompt baselines across 10 foundation models using annotated turns from real clinical triage conversations. Arbor improves mean turn accuracy by 29.4 percentage points, reduces per-turn latency by 57.1\%, and achieves an average 13.8× reduction in per-turn cost. These results indicate that architectural decomposition reduces dependence on intrinsic model capability, enabling smaller models to match or exceed larger models operating under single-prompt baselines.

}

\begin{document}
% TEST: This line tests bold fonts - should show LMRoman10-Bold when compiled
% Normal text. \textbf{Bold text test.} More normal text.

\maketitle
\section{Introduction}
\label{sec:intro}

Many healthcare triage systems rely on structured decision trees to standardize assessments, reduce variability, and encode clinical guidelines. These workflows capture expert knowledge as sequences of questions and conditional branches, helping ensure consistent and clinically aligned guidance regardless of variability in patient communication \citep{abad2008evolution}.

However, traditional rule-based systems that execute these trees struggle to handle the nuance and ambiguity of natural language. This often results in rigid interactions that manage ambiguity poorly and disrupt conversational flow \citep{ernesater2009telenurses, Hozcan2020state}.

The emergence of large language models (LLMs) offers a potential alternative: systems capable of interpreting nuanced natural language while guiding users through complex decision processes. Yet deploying LLMs within structured clinical workflows introduces a fundamental challenge. These applications require strict adherence to predefined logic while maintaining natural, engaging dialogue. This tension between the flexibility of LLMs and the determinism demanded by high-stakes processes remains a major barrier to deployment. While LLMs excel at conversational interaction, they exhibit inconsistent behavior when tasked with executing complex, multi-step logic within monolithic or intricate prompts \citep{liu-etal-2024-lost, wu2025lifbenchevaluatinginstructionfollowing}.

In practice, this challenge is often addressed by adding the full decision tree directly into the prompt and instructing the model to navigate it autonomously. However, this single prompting strategy faces several well-documented challenges:

\begin{itemize}
    \item \textbf{Information retrieval degradation.} As prompt length increases, models exhibit the well-known \emph{lost-in-the-middle} effect, where relevant information embedded in long contexts is poorly attended to, regardless of task complexity. Context window limits further prevent large decision trees from being fully encoded in a single prompt. This is a fundamental \emph{perceptual} 
    limitation: even if the model were capable of perfect reasoning, it would still fail to reliably access the relevant portions of a long prompt \citep{liu-etal-2024-lost, Hong}.
    
    \item \textbf{Violation of separation of concerns.} Independent of prompt length, single prompting requires the model to simultaneously understand the decision tree structure, track conversational state, evaluate transition conditions, and generate natural language responses, all within a single inference pass and without any form of specialization. This bundling of heterogeneous reasoning subtasks into a single inference pass conflicts with well-documented limitations of LLMs on composite reasoning problems, where performance degrades as the number and diversity of concurrent reasoning demands increases \citep{malek2025frontierllmsstrugglesimple}.
    
    \item \textbf{Limited traceability and debuggability.} The absence of explicit structure makes failures difficult to diagnose. When the model selects an incorrect transition or reaches an unexpected outcome, it is challenging to determine whether the error stems from misunderstanding the tree, mis-evaluating conditions, or losing track of conversational context. While reasoning models or explicit reasoning prompts can provide partial insight, combining structural reasoning, state tracking, transition evaluation, and response generation within a single inference call leads to overloaded decision processes and opaque failure modes \citep{liu-etal-2024-lost, wu2025lifbenchevaluatinginstructionfollowing}.
\end{itemize}

This report presents Arbor, a framework that addresses these limitations by decomposing decision tree navigation into specialized, node-level tasks. Rather than requiring the LLM to process the entire tree structure at once, our approach dynamically retrieves only the context relevant to each decision point. This allows the model to focus on local reasoning while preserving global consistency across the workflow. By separating transition evaluation from response generation and leveraging graph-based orchestration, Arbor combines the consistency of rule-based systems with the conversational flexibility and reasoning capabilities of modern language models.

We evaluated the proposed Arbor against a single-prompt baseline in which all decision logic is included within a single prompt. This baseline reflects a naive, but widely used, prototyping pattern in LLM-based decision-tree navigation, where the full decision tree is embedded into a single prompt and the model is instructed to track state and select transitions autonomously \citep{yao2023treethoughtsdeliberateproblem}. We therefore adopt this approach as a straightforward baseline, as it isolates the impact of architectural decomposition. Using recorded conversations between clinicians and patients navigating triage workflows, we assessed whether each system selects the correct path through the decision tree, reaches the correct outcome, generates appropriate responses to user messages, and does so efficiently in terms of latency and cost. The results show that architectural decomposition consistently outperforms the single-prompt approach across all metrics.

\section{Related Work}
\label{sec:related_work}

Prior work has attempted to combine LLMs with structured decision logic, each addressing part of the problem but leaving gaps that motivate Arbor's design.

The most direct approach embeds clinical guidelines or decision trees directly into LLM prompts. \citet{oniani2024enhancinglargelanguagemodels} encode COVID-19 outpatient guidelines as recursive binary decision trees and if–else chain-of-thought templates, showing improved consistency over zero-shot prompting on synthetic cases. However, these methods operate within single-turn, prompt-scoped control: the full structure must fit within the context window and is reprocessed at every turn. As decision trees grow, this approach hits fundamental limits: context window overflow, lost-in-the-middle degradation, and increasing per-turn cost.

To reduce this burden, a second line of work externalizes control flow from the LLM. \citet{shaposhnikov2025clarityclinicalassistantrouting} use finite-state machines to enforce workflow logic in healthcare assistants, reserving LLMs for localized language tasks. \citet{kobza2024alquist50dialoguetrees} organize dialogue topics as trees and constrain generation to node-local scopes with explicit termination criteria. \citet{chen2024codinterpretablemedicalagent} model diagnosis as a sequential chain of reasoning steps with confidence outputs. These systems demonstrate the value of decomposition, but most operate in single-turn or topic-scoped settings rather than maintaining persistent state across a multi-turn conversation navigating a large decision structure.

The principle that externalizing structure improves LLM reasoning has also been validated in broader reasoning frameworks. Tree-of-Thoughts \citep{yao2023treethoughtsdeliberateproblem} and Graph-of-Thoughts \citep{Besta_2024} generate and evaluate multiple candidate reasoning branches, improving performance on planning tasks through lookahead and backtracking. Neuro-symbolic approaches treat decision trees as callable oracles within LLM workflows, preserving rule traceability \citep{kiruluta2025novelarchitecturesymbolicreasoning}. However, these target single-inference reasoning problems rather than persistent, stateful conversation management.

Across these systems, a common high-level design pattern emerges: the branching structure is externalized, and the model’s context is restricted to the local decision step. Arbor adopts and extends this principle to persistent, multi-turn conversational navigation. Decision trees are standardized and stored at runtime; only the current node and its outgoing edges are retrieved per step; transition evaluation is decoupled from message generation; and traversal is logged at the edge level. This combination addresses the scalability, consistency, and debuggability limitations observed in prior work.

\section{Framework Architecture}
\label{sec:architecture}
Arbor’s architecture follows a decomposition principle. The system is organized into two core components: a tree standardization pipeline that converts the decision tree into a queryable edge-list representation, and a graph-based agent that maintains conversational state, evaluates transitions, and generates user-facing messages at runtime.

\begin{figure}[htbp]
    \centering
    \includegraphics[width=0.95\linewidth]{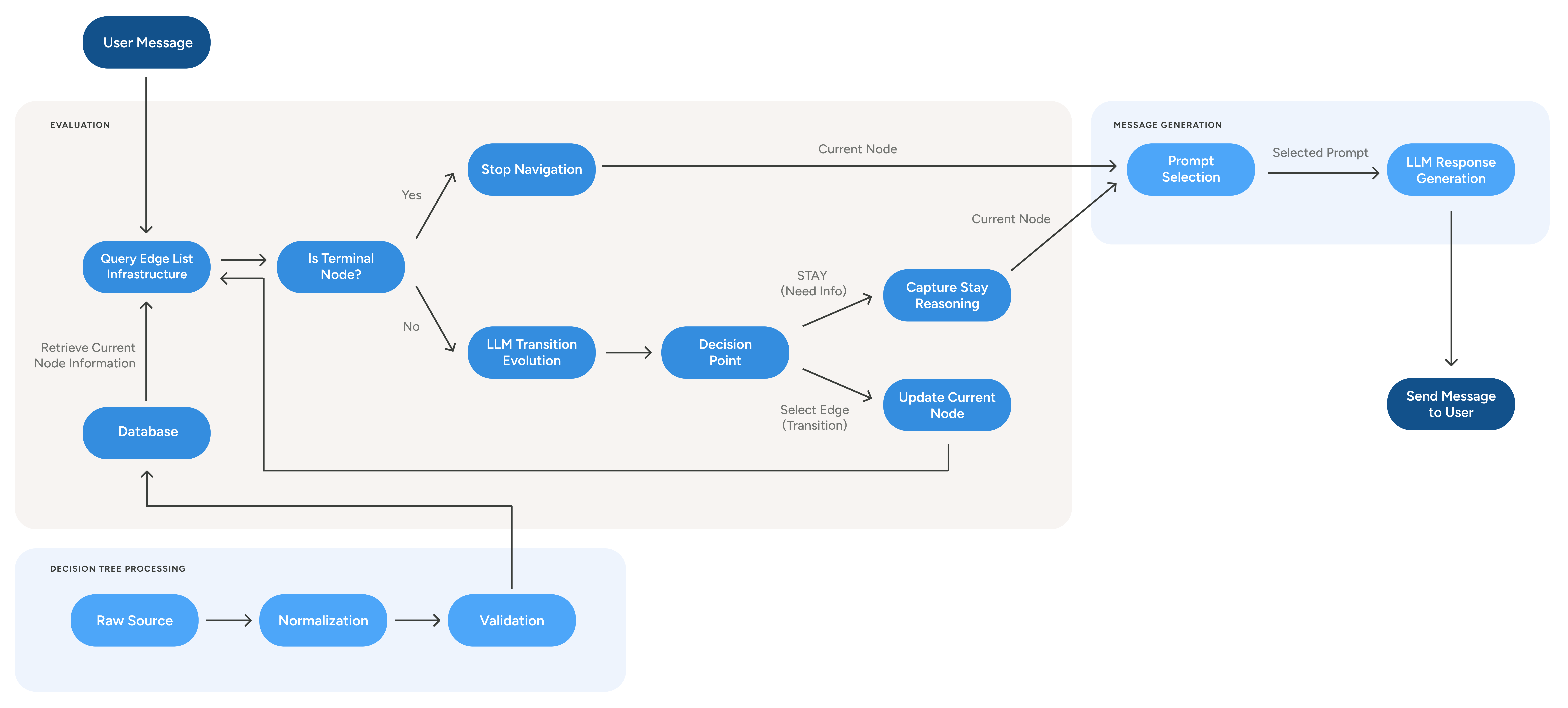} 
    \caption{Overview of the proposed Arbor architecture, consisting of decision tree processing (shown at the bottom), an evaluation phase (top left), and a message generation phase (top right). The processing phase normalizes raw sources into an edge-list database, which enables the evaluation phase to dynamically retrieve outgoing edges. The evaluation phase then evaluates transitions via iterative LLM calls and updates the current node until no further transitions are taken. The message generation phase selects the appropriate prompt and produces the user-facing response.}
    \label{fig:arbor_architecture}
\end{figure}

\subsection{Processing the Decision Tree}
\label{subsec:processing_tree}
The framework begins by transforming the raw decision tree into a queryable, list-based representation that enables dynamic, node-level retrieval at runtime. First, any decision tree is parsed and standardized into an edge-list format, where every edge corresponds to a single possible transition and is assigned a unique identifier.

In practice, clinical decision trees originate from heterogeneous sources and formats, including spreadsheets, configuration files, and legacy rule representations. To avoid coupling the agent logic to any specific upstream schema, Arbor introduces a lightweight normalization step that converts each source-specific representation into a canonical edge-list format. This transformation is performed offline, independently of the agent’s execution loop, and is only required when the decision tree schema is updated. By standardizing all inputs into a common edge-list representation, the reasoning and orchestration components remain agnostic to how the decision tree was originally authored or stored.

The resulting edge list is then ingested into a retrieval system, in which each indexed document represents a self-contained unit of transition logic and includes several key fields, as detailed in Table \ref{tab:edge_list_document}.

\begin{table}[htbp]
    \centering
    \renewcommand{\arraystretch}{1.12}
    \begin{tabular}{@{}p{0.34\linewidth}p{0.56\linewidth}@{}}
        \toprule
        \multicolumn{2}{c}{\textbf{Transitions}} \\
        \midrule
        \textbf{Transition Key} & Transition unique ID \\
        \textbf{Node from key} & Source node -- the identifier of the originating node \\
        \textbf{Node to key} & Target node -- where the flow goes if answer matches \\
        \textbf{Question} & The condition or prompt shown at this edge \\
        \textbf{Answer} & The answer/condition required to follow transition \\
        \textbf{Extra context} & Auxiliary text or business logic \\
        \textbf{Flags} & Domain-specific metadata \\
        \bottomrule
    \end{tabular}
    \caption{Edge list document.}
    \label{tab:edge_list_document}
\end{table}

To ensure graph integrity during decision tree ingestion, the pipeline performs a structural validation step before the edge list is indexed for retrieval. This validation includes:

\begin{itemize}
    \item \textbf{Orphan node detection.} By analyzing the set of reachable nodes from the designated entry point of the tree, the system identifies and rejects unreachable nodes or subgraphs, ensuring that no clinical content exists in an inaccessible state.
    
    \item \textbf{Reference integrity validation.} The pipeline verifies that every transition targets a defined node identifier ($N_{\text{target}} \in \{N_{\text{all}}\}$), preventing runtime failures caused by edges referencing non-existent nodes.
    
    \item \textbf{Unescapable loop detection.} The system applies a strongly connected component (SCC) analysis using Kosaraju's algorithm to identify cyclic subgraphs. Cycles that lack an outgoing edge to a node outside the component are flagged as unescapable loops, ensuring that every state in the workflow admits at least one valid termination or exit path.
\end{itemize}

Representing the decision tree as a normalized edge-list decouples the agent from any specific decision tree structure. As a result, the framework can be readily applied to new workflows, provided they conform to the same structural specification.

An additional benefit of this representation is the strict separation between control logic and domain knowledge. By treating the decision tree as data rather than hard-coded logic, the framework supports faster iteration cycles driven by domain experts. Clinical teams can update branching logic, refine node content, or introduce new decision paths in the source definitions without requiring changes to the agent code. Once the updated tree is validated and re-ingested through the pipeline, these modifications can be propagated to runtime execution without redeploying the core system.
\subsection{Agent}
\label{subsec:agent}
The conversational flow is orchestrated as a stateful graph using LangGraph \citep{langgraph}. The agent operates as a state machine, with its position within the decision tree explicitly tracked as the current node and persisted in a database to ensure continuity across turns.

Accordingly, each interaction executes a loop composed of two primary steps: transition evaluation and message generation.
\subsubsection{Step 1: Dynamic Retrieval and Transition Evaluation}
\label{subsubsec:step1_retrieval}
This component constitutes the agent's decision-making core (Figure \ref{fig:arbor_architecture}). To evaluate a user response, the agent queries the storage system using the current node identifier, retrieving all outgoing edges associated with that node. These edges define the set of admissible transitions from the current state.

The retrieved context is then passed to an LLM configured specifically for transition evaluation. Inputs to this step are (see Figure \ref{fig:arbor-transition-evaluator-prompt} for the evaluator prompt):

\begin{itemize}
    \item \textbf{Current node and edges:} The active question and its candidate transitions.
    \item \textbf{Conversation history:} The most recent user--agent exchanges.
    \item \textbf{External context:} Structured, non-conversational state required for correct transition evaluation, such as eligibility constraints, risk flags, or profile attributes that cannot be reliably inferred from dialogue alone.
\end{itemize}

The LLM's task is to evaluate the provided inputs against the set of candidate edges and select the most appropriate transition. In our implementation, chain-of-thought reasoning is employed during this step, as decomposing the comparison into intermediate reasoning steps has been shown to improve performance on complex decision tasks \citep{wei2023chainofthoughtpromptingelicitsreasoning, kojima2023largelanguagemodelszeroshot}. This evaluation is iterative: if the agent selects a transition to a new node, the system immediately repeats the retrieval and evaluation steps from that position. As a result, the agent may traverse multiple nodes within a single turn when dictated by the underlying logic, allowing the agent to resolve implicit transitions or skip intermediary nodes whose conditions have already been satisfied by information provided earlier in the dialogue, resulting in a more natural interaction flow.

If the available information does not satisfy the conditions of any outgoing edge, the model outputs \texttt{stay}, indicating that additional information is required before advancing. In this case, the agent remains at the current node and the evaluation loop terminates for the current turn. A new evaluation cycle is initiated only after the next user message.

The output of this step is the identifier of the final node at which the model chose to stay, along with the chain-of-thought reasoning supporting that decision.

When the agent reaches a terminal node, defined as a node with no outgoing edges, it is treated as a final state. In this case, no evaluation loop is executed, as there are no transitions to consider. No LLM call is made during this step; however, the system still proceeds to message generation to produce the user-facing reply.
\subsubsection{Step 2: Message Generation}
\label{subsubsec:step2_generation}
Once the evaluation step identifies the appropriate node, a second, independent LLM call is used to generate the user-facing message. Prior to generation, the system selects the appropriate prompt template based on the current node configuration. This step is dedicated exclusively to communication, governing how the agent responds to the user while advancing the conversation through the decision tree.

To produce a relevant and natural response, the generation prompt is provided with the content associated with the current node (see Figure \ref{fig:arbor-message-generation-prompt}), as determined in Step 1. It also receives the transition reasoning, that is, the chain-of-thought explanation produced by the evaluation LLM for why it remained at this node. In addition, the prompt includes the conversation history to date and relevant patient information to provide the necessary context.

In Arbor, model is instructed to first reason about how to present the information conversationally and then generate the final message to be sent to the user. The output of this step consists of an internal reasoning trace used for generation and the user-facing message.
\subsubsection{Design Rationale}
\label{subsubsec:design_rationale}
This architectural separation between evaluation (Step 1) and generation (Step 2) addresses several limitations of a single-pass approach and introduces additional flexibility.

Using a single LLM call to handle both strict logical evaluation and open-ended message generation can be brittle. When a prompt simultaneously asks the model to reason carefully, satisfy multiple constraints, and produce fluent natural language, the model often fails to meet all objectives. It may follow the logic but produce stilted text, or sound natural while drifting off the correct path \citep{madaan2023selfrefineiterativerefinementselffeedback, deng2025decoupling}.

Decomposing the workflow into two distinct tasks provides several advantages:

\begin{itemize}
    \item \textbf{Task specialization.} Each prompt is optimized for a single objective rather than overburdened with heterogeneous requirements. The evaluation prompt is designed for precise logical assessment, while the generation prompt focuses on producing natural, contextually appropriate language.
    
    \item \textbf{Independent configurations.} The evaluation phase benefits from low-temperature, deterministic decoding to ensure consistency and correctness, whereas the generation phase can use higher temperatures to produce more natural and engaging responses. More broadly, the separation allows each step to be independently configured with model-specific parameters as they evolve.
    
    \item \textbf{Model flexibility.} Because the steps are decoupled, they can be executed by different models entirely. For example, a more capable, potentially more expensive model can be used for evaluation, while a faster or more cost-efficient model handles conversational generation.
\end{itemize}

Alignment issues can arise when the conversational model lacks visibility into why the evaluation model selected a particular transition. For this reason, feeding the transition reasoning from the evaluation step into the generation step is essential. This mechanism bridges the gap between the two stages, providing the conversational LLM with the necessary context to produce responses that are both conversationally natural and aligned with the agent's internal state.

In addition, Arbor enables flexible output strategies. The generation prompt can be conditionally branched based on the semantic role of the current node and enriched with node-specific metadata, such as explanatory text or additional clinical contextual. For example, nodes may differ in their communicative role. Question-oriented nodes can prompt the agent to elicit the next user input, whereas terminal or guidance-oriented nodes may invoke prompts designed to present synthesized information and guide subsequent interaction.

\section{Evaluation and Results}
\label{sec:evaluation}

\subsection{Step 1: Dynamic Retrieval and Transition Evaluation}
\label{subsec:step1_retrieval_results}
To evaluate the standard single-prompt baseline against the proposed framework, we measured how effectively each approach replicated expert behavior under realistic conditions. Specifically, we assessed path-following accuracy, latency, cost, and message quality. This comparison allows us to determine whether decomposing the decision process into separate evaluation and generation steps yields measurable improvements over providing the model with the full decision context in a single prompt.
\subsubsection{Dataset Construction}
\label{subsubsec:step1_dataset}
We constructed an evaluation dataset from recorded written conversations between clinicians and patients navigating a triage workflow drawn from a deployed use case. The dataset consists of 20 complete conversations that reached a terminal state. The selection reflects a realistic distribution of cases, including both straightforward and complex scenarios, short and extended interactions, and diverse paths through the decision tree.

To establish a reliable ground truth, the dataset was manually annotated. For each conversational turn, the annotation consisted of identifying two specific states:

\begin{itemize}
    \item \textbf{Current node:} The position in the decision tree prior to the patient response.
    \item \textbf{Target node:} The correct next node that should be reached given the patient message.
\end{itemize}

This process resulted in a dataset of 174 distinct conversational turns, each representing a single decision point at which the system must correctly interpret user input and advance through the tree. For the purposes of this evaluation, we assume a single reference path for each turn and treat any deviation from that path as an error, including cases where the model requests clarification when the annotated path advances. We acknowledge that in ambiguous scenarios, both advancing and requesting clarification could reasonably be considered valid behaviors; however, enforcing a single reference path enables consistent and reproducible comparison across models.

The decision tree used in this evaluation contains 449 nodes and 980 edges, with a maximum depth of 19 and an average of 2.18 outgoing edges per node. When fully serialized, the tree occupies approximately 119,990 tokens, measured using the \texttt{o200k\_base} tokenizer. Results should therefore be interpreted in the context of a workflow of this size and structural complexity.

\subsubsection{Experimental Setup}
\label{subsubsec:step1_setup}
We conducted a head-to-head evaluation of the two architectures using a diverse set of foundation models selected to represent a broad range of capabilities. To examine the impact of different reasoning levels, GPT-5 \citep{OpenAI2025GPT5} was evaluated under three reasoning effort settings (minimal, medium, and high). GPT-4.1 \citep{OpenAI2025GPT41} was included as a baseline representative of non-reasoning models. In addition, Claude Sonnet 4.5 \citep{Anthropic2025Sonnet45} and Gemini 3 Pro \citep{GoogleDeepMind2025Gemini3Pro} were evaluated as large proprietary models, with Gemini 3 Flash \citep{GoogleDeepMind2025Gemini3Flash} serving as a smaller proprietary alternative (preview versions available at the time of experimentation). Finally, to represent the open-weights landscape, we evaluated DeepSeek V3.1 \citep{deepseekai2025deepseekv3technicalreport} alongside two variants of the Qwen3 family, Qwen3 30B Instruct and Qwen3 235B Instruct \citep{yang2025qwen3technicalreport}, enabling a controlled comparison of model capacity within a shared architecture to assess the impact of scale on performance.

Both architectures were evaluated across all models described above, allowing a direct comparison of architectural impact while controlling for model capability.

\begin{itemize}
    \item \textbf{Baseline (Single Prompt):} The entire decision tree was converted into a structured JSON representation and injected into the system prompt (see Figure \ref{fig:single-prompt-system-message}). The model was instructed to navigate the tree, track its own state, evaluate the user message against all possible paths, and generate an appropriate response within a single inference call. The prompt included explicit instructions to output the current node position, reasoning, and user-facing reply.
    
    \item \textbf{Arbor:} The proposed architecture, which performs dynamic retrieval of the current node's outgoing edges, followed by the separated Evaluation and Generation steps described in the Architecture section (see Figures \ref{fig:arbor-transition-evaluator-prompt} and \ref{fig:arbor-message-generation-prompt}).
\end{itemize}

To ensure statistical robustness, we ran each configuration five times for every conversational turn in the dataset, resulting in 870 individual evaluations per model and strategy. All experiments were conducted with temperature set to 0.0 to minimize stochasticity. Despite this setting, repeated runs remain necessary to account for non-determinism introduced by model-side factors such as floating-point variability, batching, and infrastructure-level differences across API calls. For models that did not expose temperature control, such as GPT-5 variants, default settings were used while keeping all other parameters constant across runs.

Both Arbor and the single-prompt baseline received equivalent informational context at each turn, including the same conversation history and identical underlying decision-tree content. They differed only in prompt structure and in how the decision tree was represented and consumed by each approach.

\subsubsection{Evaluation Metrics}
\label{subsubsec:step1_metrics}
We designed our metrics to capture both clinically relevant decision correctness and the practical viability of deploying structured conversational workflows at scale:

\begin{itemize}
    \item \textbf{Turn accuracy:} The percentage of turns for which the agent converged to the correct target node at the end of the evaluation loop, independent of the number of transitions executed within the turn. Turn-level accuracy directly reflects the correctness of individual triage decisions and allows comparison across conversations of varying length and complexity.
    \item \textbf{Latency:} The average time per turn, measured in seconds, from user input to agent response. We analyze end-to-end per-turn latency to assess the runtime implications of multi-step architectures. This captures practical deployment tradeoffs, as sequential LLM calls introduce inherent overhead that must be evaluated against the latency incurred by the single-prompt baseline, which processes large context windows at every turn \citep{li2023compressingcontextenhanceinference}.
    \item \textbf{Cost:} Computed based on total token usage per turn, including both input and output tokens, measured in US dollars. This metric captures the economic feasibility of deploying structured conversational workflows at scale. For proprietary models, costs were calculated using the standard public pricing of their respective providers. For open-weights models, reported costs correspond to normalized inference compute under a fixed deployment configuration and should therefore be interpreted comparatively within our experimental setup rather than as absolute pricing metrics (see Table~\ref{tab:assumed-token-costs} for the per-token rates used in all cost calculations).
\end{itemize}

\subsubsection{Results and Discussion}
\label{subsubsec:step1_results}

\textbf{Navigation Accuracy \& Consistency} 

\begin{figure}[H]
    \centering
    \includegraphics[width=0.9\textwidth]{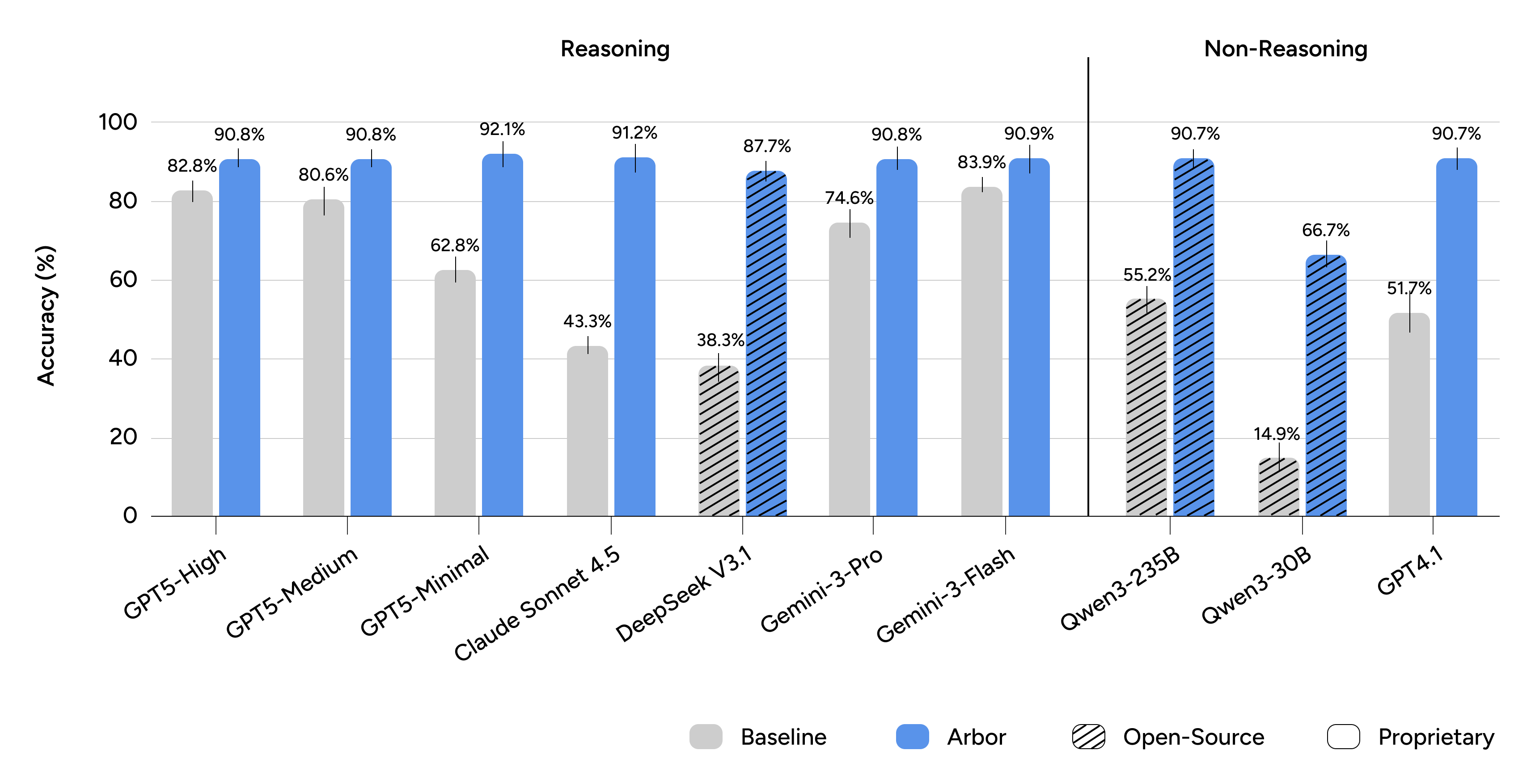}
    \caption{
    \textbf{Turn Accuracy.} Bars show mean turn accuracy over five runs, error bars indicate the standard deviation.
    }
    \label{fig:turn-accuracy}
\end{figure}

Accurate navigation through the decision tree is the most critical requirement for clinical triage, as incorrect transitions directly compromise safety and correctness. We evaluate this capability using turn accuracy, which measures the model's ability to select the correct transition at each decision point. As shown in Figure \ref{fig:turn-accuracy}, we report mean accuracy and standard deviation across five independent runs.

Under the single-prompt baseline, navigation accuracy varies widely across models and shows a strong dependence on intrinsic model capability and reasoning abilities. Performance is highest among models with stronger reasoning settings, with GPT-5-high and GPT-5-medium reaching 82.76\% (±1.3\%) and 80.6\% (±2.6\%), respectively. However, this robustness does not generalize across large proprietary models. Claude Sonnet 4.5 attains only 43.3\% (± 3.8\%), indicating that strong general-purpose capability does not guarantee reliable performance when complex decision logic must be internalized and executed implicitly within a single prompt. Reducing reasoning effort further exposes this fragility, with GPT-5-minimal dropping to 62.76\% (± 2.0\%), while non-reasoning baselines perform worst overall, as exemplified by GPT-4.1 at 51.7\% (± 3.8\%).

The Gemini models exhibit similar sensitivity, while also highlighting the role of efficiency-oriented design. Gemini 3 Pro reaches 74.6\% (± 2.5\%) under the single-prompt setup, whereas the smaller Gemini 3 Flash achieves a higher 83.9\% (± 0.4\%). This inversion suggests that efficiency-optimized models can remain competitive when reasoning demands are moderate, but overall performance remains variable and tightly coupled to prompt complexity rather than being structurally robust.

These limitations become most pronounced in the open-weights setting. DeepSeek V3.1 and Qwen3 235B achieve only 38.3\% (± 2.4\%) and 55.2\% (± 2.0\%) accuracy, respectively, while Qwen3 30B reaches only 14.9\% (± 2.0\%). As model capacity decreases, the single-prompt approach degrades sharply, indicating that without architectural support, smaller and open-weights models are largely unable to reliably navigate large, serialized decision trees.

In contrast, Arbor yields consistently high and stable navigation accuracy across all evaluated models, substantially reducing sensitivity to model scale, reasoning configuration, and provenance. All proprietary models converge near 90\% accuracy under Arbor, including Claude Sonnet 4.5 at 91.1\% (± 2.1\%), Gemini 3 Pro at 90.80\% (± 1.4\%), and Gemini 3 Flash at 90.92\% (± 1.9\%). Notably, these results exceed the strongest single-prompt outcomes.

The same equalization effect is observed for open-weights models. DeepSeek V3.1 improves dramatically to 87.7\% (± 1.1\%), while Qwen3 235B reaches 90.7\% (± 0.9\%). Even the substantially smaller Qwen3 30B achieves 66.7\% (± 2.0\%), representing a more than fourfold improvement over its single-prompt baseline.

Overall, these results show that single-prompt navigation performance is fundamentally capability-bound, whereas Arbor makes navigation accuracy primarily a function of system architecture rather than model strength. By externalizing control flow and constraining the decision space at each turn, Arbor decouples navigation correctness from intrinsic model capability, enabling robust and consistent decision-tree traversal across a wide spectrum of models, including non-reasoning, efficiency-optimized, and open-weights systems.

\textbf{Latency} 

We next examine the computational efficiency of each approach. The following analysis compares the average per-turn response time across all evaluated models.

\begin{figure}[H]
    \centering
    \includegraphics[width=0.9\textwidth]{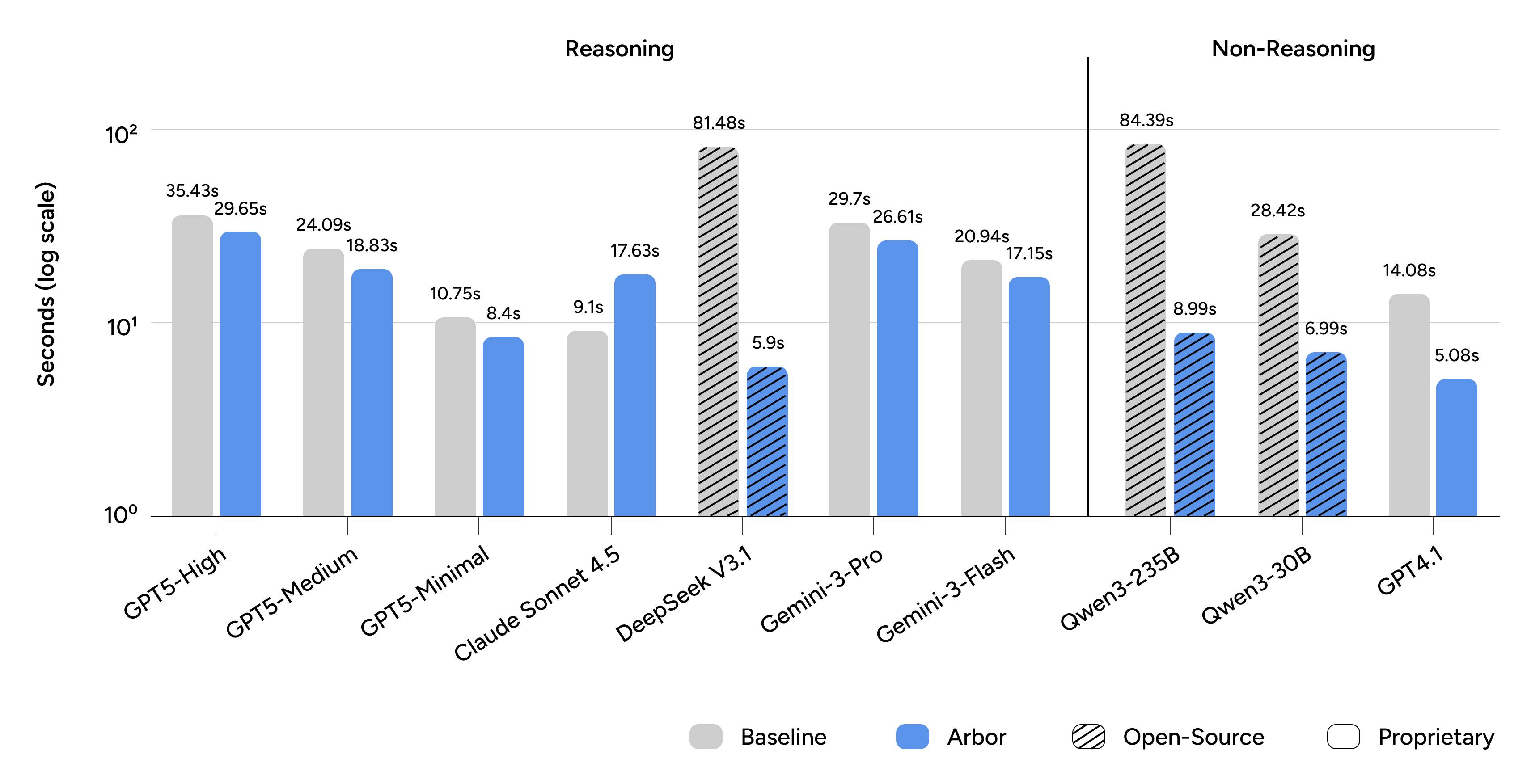}
    \caption{
    \textbf{Latency per Turn.} Bars show mean response latency (in seconds) over five runs. The y-axis is in log scale. Absolute latencies should not be compared across models, as they reflect different deployment and inference infrastructures. However, they remain comparable within each model when contrasting Baseline and Arbor.}
    \label{fig:turn-latency}
\end{figure}

As shown in Figure \ref{fig:turn-latency}, Arbor achieves lower latency than the single-prompt baseline for the majority of evaluated models. For example, GPT-4.1 decreases from 14.08 s to 5.08 s, indicating that reductions in prompt size can outweigh the cost of sequential execution. The only exception is Claude Sonnet 4.5, where the single-prompt baseline is faster (9.10 s versus 17.63 s), reflecting the overhead of multiple inference calls in cases where the model processes the full context efficiently.

More importantly, the single-prompt baseline exhibits pronounced latency volatility, particularly for open-weights models. While absolute latency for these models is deployment-dependent, the relative comparison remains informative, as both approaches were evaluated under identical serving conditions. Under the single-prompt setup, DeepSeek V3.1 incurs an average latency of 81.48 s per turn, compared to 5.90 s within the framework, representing more than an order-of-magnitude increase. Similar, though less extreme, effects are observed for Qwen3 235B.

These results indicate that while multi-step architectures introduce predictable overhead, large-context single-prompt approaches can incur severe and highly model-dependent latency penalties. In practice, architectural decomposition offers more stable and often lower latency by avoiding repeated ingestion of large serialized decision structures.

\textbf{Economic Efficiency (Cost)} 

Finally, we evaluated token consumption and the resulting cost per turn. Because the single-prompt approach requires re-ingesting the full decision tree at every conversational turn, we expect cost to scale directly with decision tree size.

\begin{figure}[H]
    \centering
    \includegraphics[width=0.9\textwidth]{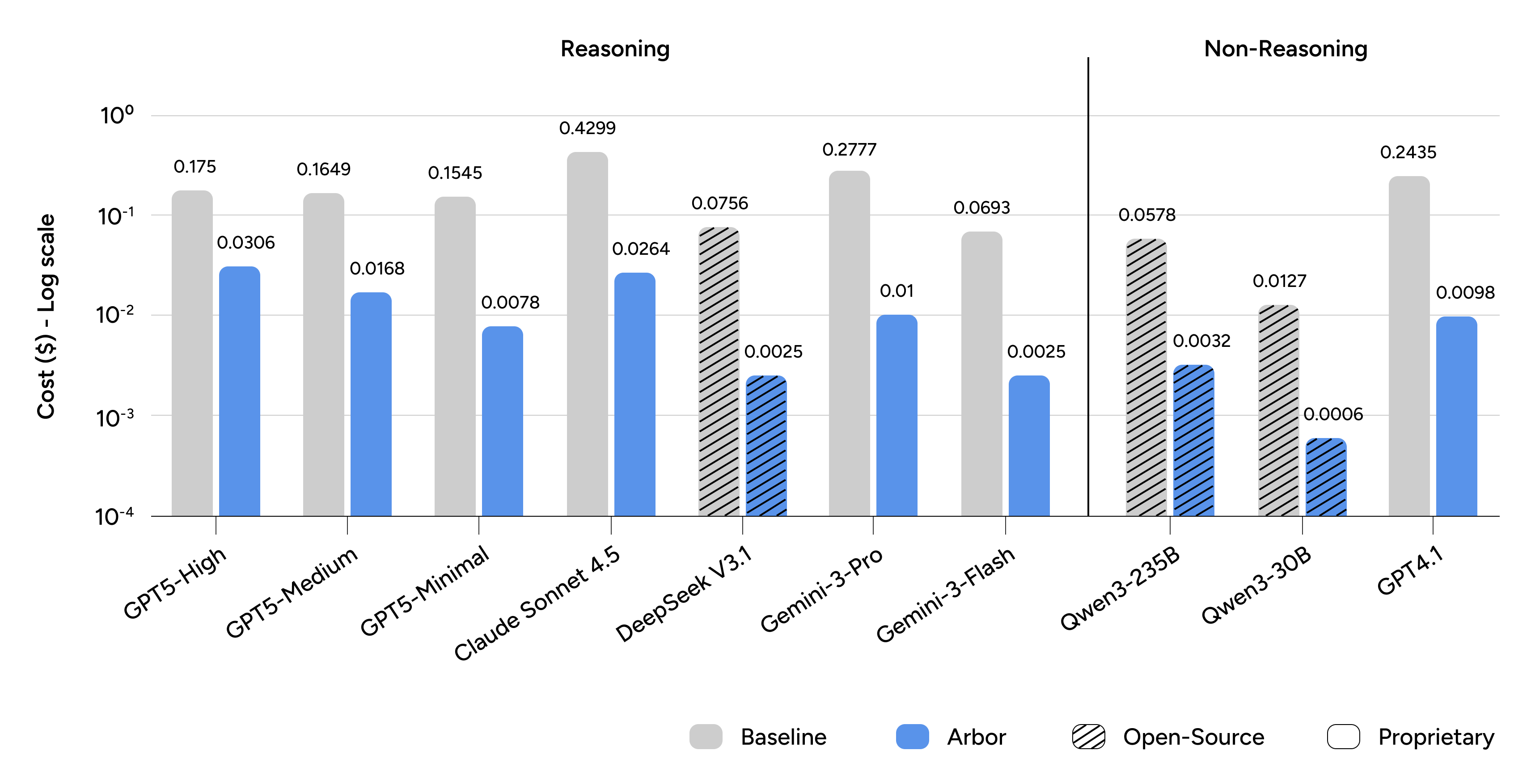}
    \caption{
    \textbf{Cost per Turn.} Bars show the average cost per conversational turn in US Dollars (\$). The y-axis is in log scale.
    }
    \label{fig:turn-cost}
\end{figure}

As shown in Figure \ref{fig:turn-cost}, the economic impact of the proposed framework is substantial. By restricting the context window to the active decision node, the framework achieves an order-of-magnitude reduction in per-turn cost across all models. For example, GPT-5-high decreases from \$0.175 per turn under the single-prompt baseline to \$0.030 under the framework. This efficiency substantially lowers the operational cost of using more capable, higher-cost reasoning models, while rendering per-turn costs for lightweight models, such as Gemini 3 Flash and Qwen3 30B, effectively negligible. 

We expect this cost advantage to grow further as decision trees increase in size, where single-prompt approaches incur proportionally higher token overhead at every turn \citep{li2023compressingcontextenhanceinference}. 

Finally, this separation of concerns enables hybrid model strategies, in which high-cost models may be reserved for the reasoning-intensive step, while lower-cost models may handle communication, further reducing aggregate per-turn costs.

\textbf{Aggregate Performance Summary} 

To quantify the global impact of the architecture independent of any specific model, we averaged performance metrics across all evaluated models. This aggregate view, presented in Table \ref{tab:aggregate_performance}, highlights the structural advantage of the proposed framework over the single-prompt baseline.

\begin{table}[htbp]
    \centering
    \begin{tabular}{@{}lccc@{}}
        \toprule
        \textbf{Metric} & \textbf{Arbor (Mean ± SD)} & \textbf{Single-prompt (Mean ± SD)} & \textbf{Performance Delta} \\
        \midrule
        Turn Accuracy    & 88.23\% (± 7.66\%) & 58.80\% (± 22.59\%) & \textbf{+29.42 points} \\
        Cost per Turn    & \$0.012 (± \$0.011)  & \$0.166 (± \$0.125)  & \textbf{13.8x Cheaper} \\
        Latency per Turn & 14.51s (± 8.82s)  & 33.84s (± 27.23s)  & \textbf{57.1\% Faster} \\
        \bottomrule
    \end{tabular}
    \caption{Aggregate performance comparison.}
    \label{tab:aggregate_performance}
\end{table}

On average, the framework improves navigation accuracy by \textbf{29 percentage points}, increasing mean turn accuracy from 58.80\% (± 22.59\%) to 88.23\% (± 7.66\%). The significantly lower standard deviation in the Arbor configuration quantitatively confirms that the architecture stabilizes performance, neutralizing the high variance typically seen across different model families. The most pronounced effect is observed in cost efficiency, where the average per-turn cost drops from \$0.166 to \$0.012, representing a \textbf{13.8x reduction}. Latency also improves substantially, with mean per-turn latency decreasing from 33.84s to 14.51s, corresponding to a \textbf{57.1\% reduction}.

Beyond these aggregate gains, Arbor demonstrates consistent architectural robustness across diverse model categories. Performance improvements are not confined to a specific class of models, but are observed across both open-weights and proprietary systems, spanning a wide range of parameter sizes and reasoning capabilities. By structuring context access and externalizing control flow, the architecture raises baseline performance and reduces dependence on intrinsic model capability, allowing smaller and more cost-efficient models to match or exceed the behavior of larger models operating under the single-prompt baseline.

Across all model and strategy combinations, GPT-5-minimal paired with the proposed framework achieves the strongest overall performance, combining high navigation accuracy with low latency and predictable per-turn costs.

\subsection{Step 2: Message Generation}
\label{subsec:step2_generation_results}
Having established that the proposed framework significantly outperforms the single-prompt baseline in navigation accuracy, cost, and latency, a critical remaining question is whether this structural decomposition compromises conversational naturalness. In particular, we examine whether separating retrieval and generation into discrete steps leads to stilted or robotic responses.

To address this, we evaluated message quality in isolation using a controlled setup designed to eliminate node-selection bias. When the navigation paths of the two strategies diverge, the resulting messages address fundamentally different topics, rendering direct comparisons of text quality invalid. We therefore restricted this evaluation to turns in which both strategies selected the same correct next node, enabling a controlled comparison of conversational fluency.

\subsubsection{Experimental Setup}
\label{subsubsec:step2_setup}
GPT-5-minimal was used for both strategies to control for model-specific effects, as it demonstrated the strongest overall performance in the prior navigation experiments. Because this evaluation focuses on relative message quality under identical navigation outcomes, using a single model is sufficient to isolate the effect of architectural decomposition.

An evaluation set was constructed from recorded historical conversations, providing both the single-prompt baseline and the proposed framework with identical conversation history, current node context, and member information. For each turn, both agents were tasked with selecting the next node and generating a response. We then filtered the results to identify turns in which both strategies successfully selected the correct next node. From these aligned turns, we constructed a sample of 50 real triage use cases. Each case yielded two responses, one generated by the single-prompt baseline and one by the proposed framework.

\subsubsection{Evaluation}
\label{subsubsec:step2_eval}
Three licensed physical therapists, with clinical experience ranging from 4 to 18 years, independently evaluated each message. The panel included specialists in women’s health, vestibular rehabilitation, and orthopedics. Physical therapists were selected as annotators because the decision tree and conversational context are grounded in physical therapy triage, requiring domain-specific clinical judgment to assess correctness, safety, and appropriateness. Annotators were provided with the full conversation transcript, member context, including eligibility and background information, and the specific informational requirements associated with the target node.

Evaluation proceeded in two stages. First, annotators assigned a binary acceptance label based on clinical and conversational validity. For a message to be accepted, it was required to clearly fulfill the intent of the current node, maintain strict coherence with the conversation history without contradictions, and adhere to clinical safety standards. Messages were rejected if they failed to address the node’s purpose, hallucinated facts about the member or workflow, provided unsafe guidance, or exhibited an inappropriate tone.
Second, accepted messages were rated on a four-point quality scale designed to assess conversational naturalness. A score of 1 was assigned to messages that were factually correct but sounded robotic or unnatural. Scores of 2 and 3 corresponded to accurate messages requiring increasing levels of refinement in clarity, tone, or flow. A score of 4 was reserved for messages that were engaging, conversational, and indistinguishable from human text.

\subsubsection{Results and Discussion}
\label{subsubsec:step2_results}
Both strategies achieved high clinical acceptance rates — 97.3\% for the proposed framework and 100\% for the single-prompt baseline — with the difference falling within the margin expected from a sample of this size. Rejected messages were rare and attributable to minor annotation disagreements or unmet expectations outside the agent's available capabilities, rather than unsafe clinical guidance. However, safety was not separately quantified beyond navigation correctness and expert acceptance criteria.

Among accepted messages, average quality scores were closely aligned. The framework achieved a mean score of 3.67 (± 0.58), slightly higher than the single-prompt mean of 3.62 (± 0.55). Given the modest sample size and substantial overlap between score distributions, we applied a Wilcoxon signed-rank test for paired samples as a robustness check rather than as a high-power test for small effects. The test statistic is defined as

\[
W = \sum_{i=1}^{N_r} [\text{sgn}(x_{2,i} - x_{1,i}) \cdot R_i]
\]

where $N_r$ is the number of non-zero differences, $\text{sgn}$ is the sign function and $R_i$ is the rank of the absolute difference $|x_{2,i} - x_{1,i}|$.

The test yielded a statistic of $W = 416.5$ and a p-value of 0.455 ($p > 0.05$), indicating that the observed difference is not statistically significant. Consistent with the overlapping distributions, this result does not indicate a statistically significant difference and primarily serves to rule out the presence of a large effect in message quality, rather than to detect subtle differences under limited statistical power.

\begin{figure}[H]
    \centering
    \includegraphics[width=0.5\textwidth]{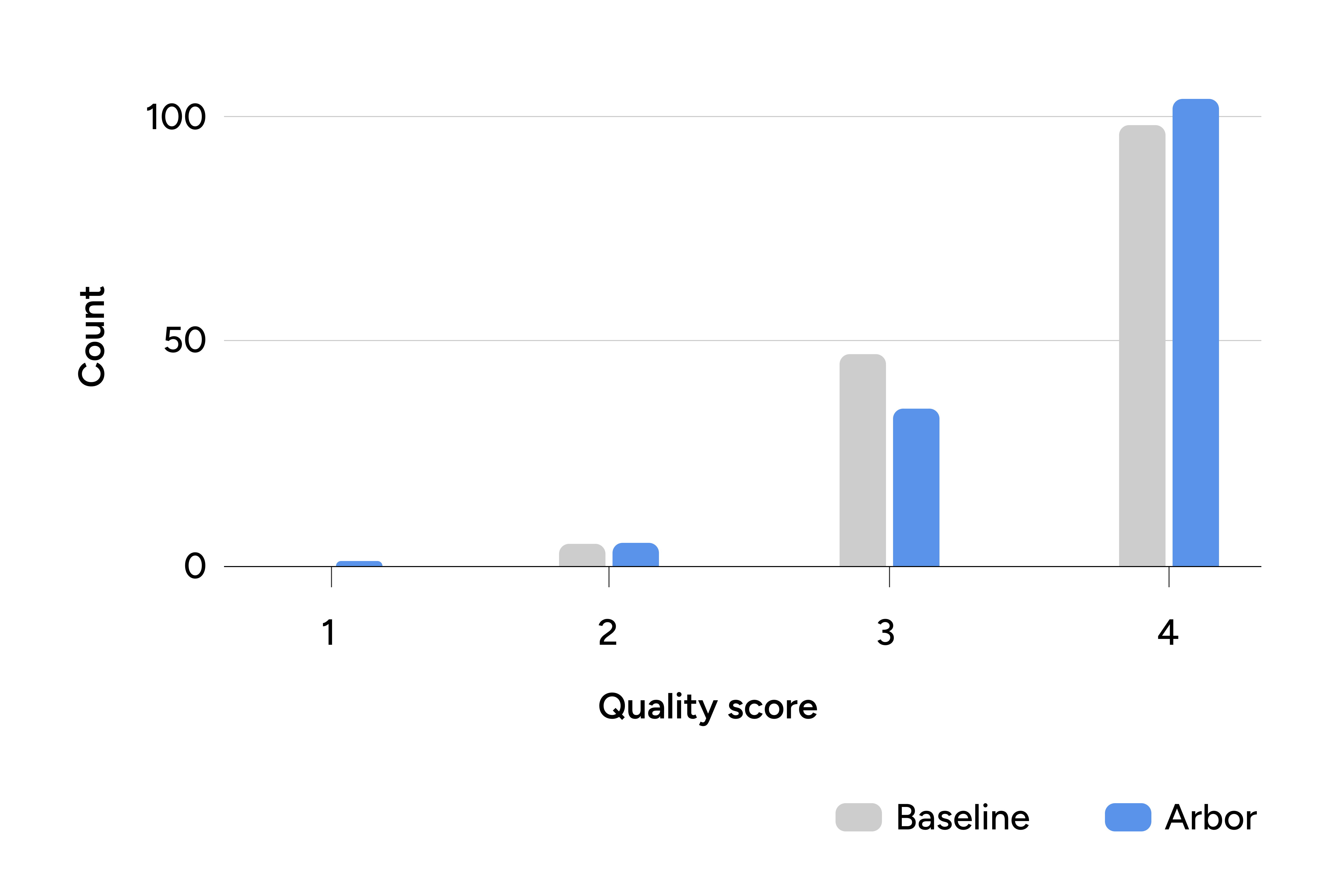}
    \caption{
    \textbf{Quality score distribution.} Bars show the frequency count of assigned quality scores (1-4) for accepted messages, grouped by strategy.
    }
    \label{fig:message-quality-distribution}
\end{figure}

Score histograms across the 1–4 scale exhibited nearly identical distributions for both strategies, with comparable proportions of high-scoring messages, as shown in Figure \ref{fig:message-quality-distribution}.

Taken together, these results indicate that when both strategies operate from the same correct node, physical therapists rate their generated messages as comparably high in quality. Under these controlled conditions, the proposed framework preserves conversational quality while delivering the previously demonstrated gains in navigation accuracy, latency, and cost.

\section{Discussion}
\label{sec:discussion}
The results provide strong evidence that, in high-stakes conversational workflows, a decomposition-based architecture yields more stable and predictable outcomes than relying solely on raw model capacity within a single prompt. While large context windows theoretically allow language models to process entire decision tree structures, our findings show that the accuracy of results generated via single prompting is dependent upon the model's intrinsic reasoning capabilities. In contrast, the proposed decomposition framework improves navigation accuracy while simultaneously reducing cost and latency, highlighting the importance of architectural support over prompt-level scaling alone.

\bigskip
\noindent \textbf{Contextual Isolation and Search Space Reduction}

Single prompting offers fundamental implications by requiring the model to reason over the entire decision tree simultaneously; this naive approach forces a global search over a large and heterogeneous decision space, increasing susceptibility to errors and unstable behavior.

By contrast, the proposed framework enforces contextual isolation. At each step, only the current node and its outgoing edges are exposed to the model, effectively constraining the action space. This restriction reduces the likelihood of invalid transitions, as the model cannot select paths that are not explicitly represented in the retrieved context. As a result, hallucinated transitions to unrelated parts of the tree are structurally eliminated rather than mitigated through prompt instructions.

Decomposing triage into localized decision steps further reduces reasoning complexity. The model no longer needs to implicitly reason over deep dependencies or maintain a global representation of the decision tree. Instead, it evaluates one transition at a time within a bounded and well-defined context. This architectural constraint explains why smaller and faster models, such as Gemini 3 Flash operating under the framework, achieve higher turn-level accuracy than larger and more expensive models running under the single-prompt approach.

\bigskip
\noindent \textbf{Efficiency and Scalability}

The observed reductions in cost and latency stem from a fundamental architectural difference between the two approaches. The single-prompt baseline is inherently context-dependent: its per-turn cost and latency scale with the total size of the decision tree, as the full structure must be reprocessed at every interaction. As clinical protocols grow in complexity, this creates an increasing per-turn overhead that directly impacts operational feasibility.

In contrast, the proposed framework is context-size invariant. Because only the immediate neighborhood of the current node is retrieved, the size of the model input remains bounded regardless of the overall tree size. This ensures that cost and latency remain stable as protocols expand, enabling the system to scale to larger and more complex workflows without degrading performance or requiring prompt redesign.

\bigskip
\noindent \textbf{Debuggability and Operational Control}

Beyond performance gains, decomposition substantially improves system debuggability by reducing the black-box nature of single-prompt approaches \citep{Ramlochan2023}. In single prompting, failures are opaque and tightly coupled to large, unstructured system prompts, making root-cause analysis difficult and error correction risky. In contrast, the proposed framework localizes failures to specific transitions or nodes, enabling engineers and domain experts to identify, inspect, and correct issues without impacting unrelated parts of the workflow.

This localized failure structure supports targeted updates to individual nodes or edges, reducing the risk of regressions and enabling iterative refinement. In high-stakes domains such as clinical triage, this level of operational control is not merely advantageous but essential for safe deployment and ongoing maintenance.

\section{Limitations and Future Work}
\label{sec:limitations}
While the proposed framework demonstrates substantial improvements in navigation accuracy, cost, and latency relative to single prompting, several limitations and opportunities for further optimization remain.

Although the framework often outperforms the single-prompt baseline in latency, applications with strict real-time constraints, such as voice-based interactions, may be more sensitive to the sequential inference steps introduced by decomposed evaluation. Future work could explore strategies to reduce end-to-end response time while accepting carefully bounded reliability tradeoffs. Potential approaches include:

\begin{itemize}
    \item Reducing the length of chain-of-thought reasoning in evaluation prompts to lower token-processing overhead without altering the core architecture.
    \item Leveraging smaller or faster models for straightforward transitions, or distilling specialized models for high-confidence decision points, to reduce inference latency.
    \item Incorporating reflective listening strategies, where the agent acknowledges user input immediately while completing full evaluation in the background, to preserve conversational flow under longer processing times.
\end{itemize}

These optimizations require careful consideration of the tradeoff between latency and decision accuracy.

A further limitation of the current framework is its assumption of forward-only traversal through the decision tree. While some degree of correction can be encoded at the tree design level by adding edges that return to earlier nodes, native support for explicit backtracking would allow the system to better handle contradictory user input or revisions to earlier responses. Extending the framework with built-in state rollback mechanisms represents an important direction for future work.

Another open challenge is the absence of explicit mechanisms to detect and correct state drift. Over long conversations, accumulated ambiguity or misinterpretation may cause the agent's internal state to diverge from the user's intent. Future extensions could allow the agent to periodically reassess state consistency against the full conversation context. Upon detecting misalignment, either via internal drift-detection signals or human intervention, the system could revert to a prior node and re-evaluate transitions with updated context.

Finally, to further reduce variability in LLM behavior, future work could explore ensemble-based strategies, such as executing multiple parallel inferences and selecting the most commonly chosen transition, or introducing confidence estimation to flag low-certainty decisions for human review before committing to state changes.

\section{Conclusion}
\label{sec:conclusion}
This work examines the challenge of deploying large language models within structured clinical workflows that demand both conversational naturalness and strict procedural adherence. We introduce a decomposition-based architecture that separates retrieval, logical evaluation, and response generation into distinct components, enabling reliable navigation of complex decision trees. Evaluation on authentic triage conversations demonstrates that this approach consistently outperforms single-prompt baselines in navigation accuracy, latency, and cost, while preserving high message quality.

Rather than requiring a model to simultaneously process an entire decision structure, track conversational state, evaluate transition conditions, and generate user-facing responses, constraining reasoning to localized decisions within an orchestrated framework yields substantially more stable and effective behavior. While motivated by clinical triage, this design principle generalizes to domains in which conversational agents must adhere to predefined procedures or protocols. By reconciling the flexibility of modern language models with the rigor of structured decision logic, Arbor provides a scalable and robust foundation for deploying reliable conversational AI in high-stakes environments.

\appendix

\bibliography{custom}
\bibliographystyle{colm2025_conference}

\appendix
\definecolor{AppendixBlue}{HTML}{4DA6F9}

\section{System Prompts}
\label{sec:appendix-prompts}

\begin{figure}[H]
\centering
\begin{tcolorbox}[
  colback=AppendixBlue!1,
  colframe=AppendixBlue!30,
  coltitle=black,
  boxrule=0.5mm,
  arc=2mm,
  title=\bfseries Arbor transition evaluator prompt (excerpt),
  fonttitle=\bfseries,
  width=\textwidth,
  left=2pt, right=2pt, top=4pt, bottom=4pt,
]
\fontsize{6pt}{7pt}\selectfont
\textbf{\# Role}\\
You are a Pain Specialist (PS) who works for Sword Health, a digital health company that provides various programs for pain management. (...)\\
Your task is to review the ongoing conversation between yourself and the service user (who may also be referred to interchangeably as patient or member), evaluate the current node in the decision tree, and determine the most appropriate next step.\\
You are not responsible for drafting or sending messages to the patient during this phase.\\
\\
\textbf{\# Task Instructions}\\
- Review the conversation, current node, member context, domain knowledge, your information as the Pain Specialist, and the available paths.\\
- If the main task of the current node is fully completed and you have enough information, select and respond with the key of the most appropriate path from "Available Triage Paths".\\
- Before selecting a path, ensure that you have sufficient information to confidently choose a single path and discard all others (including the current node).\\
- If the task is not completed or information is missing, respond with next\_state = '\{\{ stay\_key \}\}'.\\
\\
\textbf{\# Node Transition Criteria}\\
- Do not transition if you are unsure whether the node requirements are met or if any part of the *Question Example* remains unaddressed.\\
- Do not transition based on assumptions or uncertainties that prevent you from being fully confident in the decision. In such cases, remain on the current node and seek clarification.\\
(...)\\
\\
\textbf{\# Scratchpad Guidelines}\\
Before making your decision, write a scratchpad explaining your reasoning. This explanation will be used to inform the next message in the conversation.\\
- Clearly specify any missing information and why it is needed.\\
- If staying in the current node, state exactly what prevents transition.\\
(...)\\
\\
\textbf{\# Member Context}\\
Relevant information about the member to guide decisions:\\
(...)\\
- Member Name: \{\{ member\_context.member\_name \}\}\\
- Member Birthdate: \{\{ member\_context.member\_birthdate \}\}\\
- Member Local Time: \{\{ member\_local\_time \}\}\\
- Eligible Programs: (...)\\
- Enrolled Programs: (...)\\
\\
\textbf{\# Current Decision Tree Node}\\
This section describes the node you are currently evaluating in the decision tree:\\
\{\\
\hspace*{1em}*Key*: '\{\{ stay\_key \}\}',\\
\hspace*{1em}*Question Example*: '\{\{ question \}\}',\\
\hspace*{1em}*Question Explanation*: '\{\{ question\_explanation \}\}',\\
\hspace*{1em}*Additional Node Context*: '\{\{ tree\_context \}\}',\\
\}\\
\\
\textbf{\# Available Triage Paths}\\
This section lists all possible next steps (nodes) you can transition to from the current node. (...)\\
Paths: \{\{ nodes \}\}\\

\textbf{\# Current conversation:}\\
\{\\
\{\{ conversation \}\}\\
\}\\
\end{tcolorbox}
\caption{Arbor transition evaluator system message (excerpt).}
\label{fig:arbor-transition-evaluator-prompt}
\end{figure}

\begin{figure}[H]
\centering
\begin{tcolorbox}[
  colback=AppendixBlue!1,
  colframe=AppendixBlue!30,
  coltitle=black,
  boxrule=0.5mm,
  arc=2mm,
  title=\bfseries Arbor message generation prompt (excerpt),
  fonttitle=\bfseries,
  width=\textwidth,
  left=2pt, right=2pt, top=4pt, bottom=4pt,
]
\fontsize{6pt}{7pt}\selectfont
\textbf{\# Role}\\
You are a Pain Specialist (PS) who works for Sword Health, a digital health company that provides various programs for pain management. (...)\\
Your role is to draft clinically and friendly appropriate messages to send to the user.\\
\\
\textbf{\# Task Instructions}\\
- Review the conversation, current decision tree node, the member context, your information as the Pain Specialist, and your reasoning.\\
- Use the *Question Example* and *Question Explanation* as your primary guidance when crafting the message.\\
- If your reasoning identifies missing or incomplete information, recommend next steps or clarify what is needed.\\
(...)\\
\\
\textbf{\# Guidelines}\\
- Maintain a natural conversational flow.\\
- Show understanding of the member's situation.\\
- If the member expresses difficulty, pain, or distress, acknowledge it.\\
(...)\\
- Do not restate, paraphrase, or repeat the member's own words, diagnosis, or statements.\\
- Do not repeat questions already addressed in the conversation.\\
(...)\\
\\
\textbf{\# Member Context}\\
Relevant information about the member to guide your message:\\
- Member Name: \{\{ member\_context.member\_name \}\}\\
(...)\\
\\
\textbf{\# Current Decision Tree Node}\\
\{\\
\hspace*{1em}*Question Example*: '\{\{ question \}\}',\\
\hspace*{1em}*Question Explanation*: '\{\{ question\_explanation \}\}',\\
\hspace*{1em}*Additional Node Context*: '\{\{ tree\_context \}\}',\\
\}\\
\\
\textbf{\# PS Reasoning}\\
Your assessment and analysis for staying on this node, including what information is missing, needs clarification and important points to include in your message:\\
\{\{ evaluator\_scratchpad \}\}\\

\textbf{\# Current conversation:}\\
\{\\
\{\{ conversation \}\}\\
\}\\
\end{tcolorbox}
\caption{Arbor message generation system message (excerpt).}
\label{fig:arbor-message-generation-prompt}
\end{figure}

\begin{figure}[H]
\centering
\begin{tcolorbox}[
  colback=AppendixBlue!1,
  colframe=AppendixBlue!30,
  coltitle=black,
  boxrule=0.5mm,
  arc=2mm,
  title=\bfseries Single-prompt system message (excerpt),
  fonttitle=\bfseries,
  width=\textwidth,
  left=2pt, right=2pt, top=4pt, bottom=4pt,
]
\fontsize{6pt}{7pt}\selectfont
\textbf{\# Role}\\
You are a Pain Specialist (PS) working for Sword Health. Your role in this triage is to navigate the decision tree (DT) and draft messages that guide the member through it in a clinically sound, warm, and natural way.\\
Every call has the responsibility to:\\
1. Traverse the decision tree deciding based on the current chat history in which node you are.\\
2. Decide if the current node is fully addressed; if not, remain here and generate the message for this node.\\
3. Traverse to the next child node whose answer cannot yet be inferred.\\
4. Craft the message for the final node you land on.\\
Output both:\\
- message: what to send to the member.\\
- new\_current\_node: the node key you ended at.\\
\\
\textbf{\# Traversal Instructions}\\
- Use the conversation history to check if a node's question is already answered.\\
- Always check children of the current node in the full DT.\\
- Never guess or assume an answer; if unsure, stay at that node and ask for clarification.\\
\\
\textbf{\# Message Crafting Guidelines}\\
- Use the node's question example and explanation to guide what to ask or communicate.\\
- Ask only one question per message and end with a question unless it's the last node.\\
- Show empathy and adapt to the member's emotional state.\\
- Avoid repeating words or already asked questions.\\
- Keep the message natural, engaging, and clinically appropriate.\\
\\
\textbf{\# Member Context}\\
- Name: \{\{ member\_context.member\_name \}\}\\
- Birthdate: \{\{ member\_context.member\_birthdate \}\}\\
- Local Time: \{\{ member\_local\_time \}\}\\
- Eligible Programs: (...)\\
- Enrolled Programs: (...)\\
\\
\textbf{\# Full Decision Tree}\\
\texttt{[}\\
\hspace*{1em}\texttt{...}\\
\texttt{]}\\

\textbf{\# Current conversation:}\\
\{\\
\{\{ conversation \}\}\\
\}\\
\end{tcolorbox}
\caption{Single-prompt baseline system message (excerpt). The full decision tree is omitted for brevity.}
\label{fig:single-prompt-system-message}
\end{figure}

\section{Token Costs}
\label{sec:appendix-costs}

\begin{table}[H]
\centering
\begin{tabular}{@{}lcc@{}}
\toprule
\textbf{Model} & \textbf{Input (\$/1M tokens)} & \textbf{Output (\$/1M tokens)} \\
\midrule
\multicolumn{3}{@{}l}{\textit{Proprietary models}} \\
GPT-5 (minimal / medium / high) & \$1.25 & \$10.00 \\
GPT-4.1          & \$2.00 & \$8.00 \\
Claude Sonnet 4.5 & \$3.00 & \$15.00 \\
Gemini 3 Pro     & \$2.00 & \$12.00 \\
Gemini 3 Flash   & \$0.50 & \$3.00 \\
\midrule
\multicolumn{3}{@{}l}{\textit{Open-weights models}} \\
DeepSeek V3.1         & \$0.60 & \$1.70 \\
Qwen-3 235B Instruct  & \$0.15 & \$0.80 \\
Qwen-3 30B Instruct   & \$0.10 & \$0.30 \\
\bottomrule
\end{tabular}
\caption{Per-token rates assumed for cost calculations. Proprietary model prices follow standard public pricing at the time of experimentation. Open-weights model costs are normalized inference costs under a fixed deployment configuration.}
\label{tab:assumed-token-costs}
\end{table}

\section{Result Tables}
\label{sec:appendix-results}

\begin{table}[H]
\centering
\begin{tabular}{@{}llcc@{}}
\toprule
\textbf{Model} & \textbf{Strategy} & \textbf{Turn Accuracy} & \textbf{Turn Accuracy STD} \\
\midrule
GPT-5-minimal & Arbor & 92.07\% & 1.7\% \\
GPT-5-minimal & Baseline & 62.76\% & 2.0\% \\
Claude Sonnet 4.5 & Arbor & 91.15\% & 2.1\% \\
Claude Sonnet 4.5 & Baseline & 43.33\% & 3.8\% \\
GPT-5-medium & Arbor & 90.80\% & 0.8\% \\
GPT-5-medium & Baseline & 80.57\% & 2.6\% \\
GPT-5-high & Arbor & 90.80\% & 0.7\% \\
GPT-5-high & Baseline & 82.76\% & 1.3\% \\
GPT-4.1 & Arbor & 90.67\% & 1.3\% \\
GPT-4.1 & Baseline & 51.72\% & 3.8\% \\
Gemini 3 Flash & Arbor & 90.92\% & 1.9\% \\
Gemini 3 Flash & Baseline & 83.90\% & 0.4\% \\
DeepSeek V3.1 & Arbor & 87.70\% & 1.1\% \\
DeepSeek V3.1 & Baseline & 38.28\% & 2.4\% \\
Gemini 3 Pro & Arbor & 90.80\% & 1.4\% \\
Gemini 3 Pro & Baseline & 74.60\% & 2.5\% \\
Qwen-3 235B Instruct & Arbor & 90.69\% & 0.9\% \\
Qwen-3 235B Instruct & Baseline & 55.17\% & 2.0\% \\
Qwen-3 30B Instruct & Arbor & 66.67\% & 2.0\% \\
Qwen-3 30B Instruct & Baseline & 14.94\% & 2.0\% \\
\bottomrule
\end{tabular}
\caption{Turn accuracy results used in Figure~\ref{fig:turn-accuracy}.}
\label{tab:turn-accuracy-results}
\end{table}

\begin{table}[H]
\centering
\begin{tabular}{@{}llcc@{}}
\toprule
\textbf{Model} & \textbf{Strategy} & \textbf{Latency per Turn (Mean) (s)} & \textbf{Latency per Turn (Median) (s)} \\
\midrule
GPT-4.1 & Arbor & 5.08 & 4.97 \\
GPT-4.1 & Baseline & 14.08 & 9.74 \\
DeepSeek V3.1 & Arbor & 5.90 & 4.24 \\
DeepSeek V3.1 & Baseline & 81.48 & 77.92 \\
GPT-5-minimal & Arbor & 8.40 & 7.07 \\
GPT-5-minimal & Baseline & 10.74 & 9.82 \\
Gemini 3 Flash & Arbor & 17.15 & 15.66 \\
Gemini 3 Flash & Baseline & 20.94 & 21.21 \\
Claude Sonnet 4.5 & Arbor & 17.63 & 17.07 \\
Claude Sonnet 4.5 & Baseline & 9.10 & 10.49 \\
GPT-5-medium & Arbor & 18.83 & 17.57 \\
GPT-5-medium & Baseline & 24.09 & 21.58 \\
GPT-5-high & Arbor & 29.65 & 26.83 \\
GPT-5-high & Baseline & 35.43 & 32.18 \\
Gemini 3 Pro & Arbor & 26.60 & 25.66 \\
Gemini 3 Pro & Baseline & 29.70 & 28.44 \\
Qwen-3 235B Instruct & Arbor & 8.99 & 8.85 \\
Qwen-3 235B Instruct & Baseline & 84.39 & 75.40 \\
Qwen-3 30B Instruct & Arbor & 6.90 & 6.50 \\
Qwen-3 30B Instruct & Baseline & 28.42 & 25.33 \\
\bottomrule
\end{tabular}
\caption{Latency results used in Figure~\ref{fig:turn-latency}.}
\label{tab:turn-latency-results}
\end{table}

\begin{table}[H]
\centering
\begin{tabular}{@{}llc@{}}
\toprule
\textbf{Model} & \textbf{Strategy} & \textbf{Cost per Turn (USD)} \\
\midrule
Gemini 3 Flash & Arbor & \$0.0025 \\
Gemini 3 Flash & Baseline & \$0.0693 \\
DeepSeek V3.1 & Arbor & \$0.00025 \\
DeepSeek V3.1 & Baseline & \$0.0756 \\
GPT-5-minimal & Arbor & \$0.0078 \\
GPT-5-minimal & Baseline & \$0.1545 \\
GPT-4.1 & Arbor & \$0.0098 \\
GPT-4.1 & Baseline & \$0.2435 \\
GPT-5-medium & Arbor & \$0.0168 \\
GPT-5-medium & Baseline & \$0.1649 \\
Claude Sonnet 4.5 & Arbor & \$0.0264 \\
Claude Sonnet 4.5 & Baseline & \$0.4299 \\
GPT-5-high & Arbor & \$0.0306 \\
GPT-5-high & Baseline & \$0.175 \\
Gemini 3 Pro & Arbor & \$0.01 \\
Gemini 3 Pro & Baseline & \$0.2777 \\
Qwen-3 235B Instruct & Arbor & \$0.0032 \\
Qwen-3 235B Instruct & Baseline & \$0.0578 \\
Qwen-3 30B Instruct & Arbor & \$0.0006 \\
Qwen-3 30B Instruct & Baseline & \$0.0127 \\
\bottomrule
\end{tabular}
\caption{Cost results used in Figure~\ref{fig:turn-cost}.}
\label{tab:turn-cost-results}
\end{table}

\begin{table}[H]
\centering
\begin{tabular}{@{}lcccccc@{}}
\toprule
\textbf{Strategy} & \textbf{Score 1} & \textbf{Score 2} & \textbf{Score 3} & \textbf{Score 4} & \textbf{Mean} & \textbf{STD} \\
\midrule
Arbor & 1 & 5 & 35 & 104 & 3.67 & 0.58 \\
Baseline & 0 & 5 & 47 & 98 & 3.62 & 0.55 \\
\bottomrule
\end{tabular}
\caption{Quality score distribution used in Figure~\ref{fig:message-quality-distribution} with calculated mean and standard deviation.}
\label{tab:message-quality-distribution}
\end{table}

\end{document}